\documentclass[journal,letterpaper]{IEEEtran}
\usepackage[T1]{fontenc}

\usepackage{amsmath,amssymb,amsfonts}
\usepackage{mathrsfs}

\usepackage{booktabs}
\usepackage{array}
\usepackage{tabularx}
\usepackage{adjustbox}
\usepackage{placeins}
\usepackage{algorithm}
\usepackage{algorithmic}
\usepackage{threeparttable}

\usepackage{graphicx}
\usepackage{cite}
\usepackage{url}
\usepackage{textcomp}
\usepackage[final]{microtype}

\usepackage{pifont}

\newcommand{\cmark}{\ding{51}}
\newcommand{\xmark}{\ding{55}}
\newcommand{\pxmark}{\ding{51}\kern-0.45em\ding{55}}
\newcolumntype{Y}{>{\raggedright\arraybackslash}X}
\newcolumntype{L}[1]{>{\raggedright\arraybackslash}p{#1}}
\newcolumntype{C}[1]{>{\centering\arraybackslash}p{#1}}
\newcommand{\tabnoteskip}{\vspace{0.55em}}

\begin{document}

\title{SCENIC: Semantic-Conditioned Edge-Aware Neural Framework for Structured IoT Command Generation}

\author{Luke~Ztz~Hu, Hongbing~Lang, and Songping~Mai,~\IEEEmembership{Member,~IEEE}

\thanks{This work was supported by the Shenzhen Science and Technology Innovation Bureau under Grant ZDCY20250901110707008. (Corresponding author: Songping Mai; e-mail: mai.songping@sz.tsinghua.edu.cn.)}%

 \thanks{Luke Ztz Hu and Songping Mai are with the Tsinghua Shenzhen International Graduate School, Shenzhen, Guangdong, 518055 China (e-mail: zhu-tz24@mails.tsinghua.edu.cn; mai.songping@sz.tsinghua.edu.cn).}%
 \thanks{Hongbing Lang is with Shenzhen Belon Technology Co., Ltd., Shenzhen, China (e-mail: hblang@belon.cn).}%
 }

\markboth{IEEE Internet of Things Journal}%
{Hu \MakeLowercase{\textit{et al.}}: SCENIC: Semantic-Conditioned Edge-Aware Neural Framework}

\maketitle
\begin{abstract}
Edge Internet of Things (IoT) agents are often constrained by memory capacity, privacy requirements, communication latency, and recurring inference cost. Current smart-home assistants commonly rely on API-level command interfaces or cloud-based language models that remain difficult to deploy on edge devices. This paper addresses edge IoT command generation as a many-to-one structured output task, where multiple natural-language instructions map to the same canonical command string for deterministic smart-home parsing. To support this setting, we propose \textbf{S}emantic-\textbf{C}onditioned \textbf{E}dge-Aware \textbf{N}eural Framework for Structured \textbf{I}oT \textbf{C}ommand Generation (SCENIC), an end-to-end framework covering model architecture selection, Smart Home Instruct data generation, triplet-loss contrastive supervised fine-tuning, pruning and quantization, and deployment-oriented export. We evaluate sub-0.2B-scale transformer backbones, which are, to the best of our knowledge, among the smallest language-model backbones studied for edge IoT structured command generation. On Smart Home Instruct-Bench, the strongest dense decoder-only row reaches 99.0\% EM@1, while the encoder--decoder model retains stronger high-sparsity behavior. A representative pruned INT8 encoder--decoder export preserves 91.0\% EM@1 and 99.0\% EM@5 while reducing exported model size by 25.38\%. TensorRT profiling of the NVIDIA 2:4 sparse encoder export further shows up to 1.8$\times$ encoder-component speedup, indicating that the selected encoder--decoder deployment path can retain structured command accuracy under edge-oriented compression while hardware acceleration evidence remains component-level. The SCENIC code and experimental artifacts are open sourced to support reproducibility.
\end{abstract}
\section{Introduction}
\IEEEPARstart{T}{ransformer}-based large language models (LLMs) have significantly advanced natural language processing by enabling sophisticated semantic understanding and generation across diverse tasks~\cite{vaswani2017attention}. Recent progress has largely been driven by scaling model parameters, training data, and computational resources, resulting in models with billions to trillions of parameters that achieve strong emergent capabilities across reasoning and generation tasks~\cite{kaplan2020scaling}. However, these large-scale models require substantial memory, computational throughput, and cloud infrastructure, limiting their practicality for resource-constrained Internet of Things (IoT) edge environments.

For smart-home assistants, cloud-based LLM APIs can provide strong semantic reasoning and conversational performance. However, reliance on remote cloud inference introduces recurring operational costs, communication latency, and potential privacy concerns due to external transmission of user data. As a result, recent research has increasingly focused on compact language models capable of local edge deployment for smart-home and IoT applications. Maintaining performance under aggressive quantization and pruning remains challenging, particularly for compact decoder-only auto-regressive architectures operating in resource-constrained environments~\cite{zheng2025edgeLLMsurvey,li2025homebench}. Prior to the emergence of large-scale auto-regressive language models, encoder-only architectures such as BERT~\cite{devlin2019bert} and its variants dominated natural language understanding benchmarks through bidirectional contextual encoding tasks. For many-to-one structured command-generation tasks, encoder--decoder models such as T5~\cite{raffel2020t5} naturally align diverse linguistic formulations to a common structured target through cross-attention between the encoded instruction and generated output, whereas encoder-only architectures such as BERT~\cite{devlin2019bert} may similarly benefit from their bidirectional semantic representations, which have historically excelled on language-understanding tasks.

Motivated by these observations, this work investigates practical architecture and output-formulation choices for edge IoT deployment. Specifically, we evaluate two self-trained 0.2B encoder-only and decoder-only models, Encoder-Chinese-SLM~\cite{hu2026encoderchineseslm} and Decoder-Chinese-SLM~\cite{hu2026decoderchineseslm}, along with \textit{ChatLM-mini-Chinese} (CMC)~\cite{chen2023chatlmmini}, a 0.2B encoder--decoder model, under data-scarce smart-home assistant settings. We propose \textbf{S}emantic-\textbf{C}onditioned \textbf{E}dge-Aware \textbf{N}eural Framework for Structured \textbf{I}oT \textbf{C}ommand Generation (SCENIC). The SCENIC framework streamlines model selection to edge deployment for IoT assistants.
\section{Related Work}
\subsection{Edge LLMs for IoT and Smart Home Assistants}
LLM architectures have largely favored decoder-only models, including LLaMA~\cite{touvron2023llama}, Qwen~\cite{yang2025qwen25,bai2023qwen,yang2024qwen2,yang2025qwen3}, and ChatGLM~\cite{glm2024chatglm}, due to their scalability and strong open-ended instruction-following capability. However, these architectures primarily target large-scale general-purpose generation and therefore introduce substantial computational and memory overhead for smart-home edge applications. Birkmose \textit{et al.}~\cite{birkmose2025homeAssistant} investigated fully on-device smart-home assistants using compact decoder-only SmolLM2 models with 0.36B and 1.7B parameters on 8\,GB edge hardware. Their system performed intent detection, slot filling, JSON action generation, and response generation for lighting, thermostat, and device-control commands. While 8-bit and 16-bit quantization largely preserved semantic performance, aggressive 4-bit quantization reduced device-service classification accuracy under noisy real-world prompts. Similar edge-oriented LLM systems predominantly rely on decoder-only autoregressive architectures, whose decoding process introduces reduced robustness under aggressive quantization and sparsification for structured edge-generation tasks~\cite{yang2025qwen25}. Recent surveys further highlight that compact edge LLM systems remain constrained by quantization robustness, inference efficiency, and hardware-aware deployment optimization under resource-limited environments~\cite{zheng2025edgeLLMsurvey,khatiwada2025iotLLMsurvey}. Although prior edge-oriented LLM studies demonstrate the feasibility of compact on-device inference, they mainly evaluate whether small models can execute smart-home language tasks under constrained hardware. They do not directly answer how compact transformer architecture and output formulation interact with structured command accuracy after pruning and quantization. This gap is important for edge IoT deployment, where compression must reduce storage or computation without disrupting deterministic device-control behavior.
Raffel \textit{et al.}~\cite{raffel2020t5} proposed T5, a unified encoder--decoder transformer for sequence-to-sequence learning. Based on the T5 architecture, Chen~\cite{chen2023chatlmmini} developed CMC, a compact 0.2B Chinese language model designed for lightweight deployment. Devlin \textit{et al.}~\cite{devlin2019bert} provided BERT, demonstrating the effectiveness of bidirectional encoder representations for language-understanding tasks. This work revisits compact encoder--decoder and encoder-only architectures for semantic smart-home command generation, where diverse natural-language instructions correspond to deterministic device-control outputs. 
\subsection{Smart Home Instruction Benchmarking}
Recent LLM-based smart-home assistants use large cloud-hosted models as the central reasoning component, while edge devices mainly serve as sensing, communication, or command relay units. This architecture benefits from the broad instruction-following ability of frontier LLMs, but it does not fully solve low-resource edge applications.
Li \textit{et al.}~\cite{li2025homebench} proposed HomeBench, a smart-home benchmark designed to evaluate LLMs under valid and invalid instructions across both single-device and multi-device scenarios. HomeBench contains 100 smart-home scenarios and more than 170K instructions, covering five instruction types: valid single-device (VS), invalid single-device (IS), valid multi-device (VM), invalid multi-device (IM), and mixed multi-device (MM) instructions~\cite{li2025homebench}. 
\begin{table}[t]
\centering
\caption{Smart-home benchmark coverage adapted from HomeBench.}
\label{tab:homebench_taxonomy_comparison}
\scriptsize
\renewcommand{\arraystretch}{1.12}
\setlength{\tabcolsep}{2.2pt}
\begin{threeparttable}
\begin{tabular*}{\columnwidth}{@{\extracolsep{\fill}}lccccccc@{}}
\toprule
\textbf{Resource} &
\textbf{VS} &
\textbf{IS} &
\textbf{VM} &
\textbf{IM} &
\textbf{MM} &
\textbf{II} &
\textbf{Size} \\
\midrule
IFTTT~\cite{yu2021ifttt}
& \pxmark & \xmark & \pxmark & \xmark & \xmark & \xmark & 50K+ \\

SAGE~\cite{rivkin2025aiotSmartHome}
& \cmark & \xmark & \xmark & \xmark & \xmark & \xmark & 50 \\

Home Assistant Requests~\cite{acon96homeassistant2024}
& \cmark & \xmark & \xmark & \xmark & \xmark & \xmark & 30K+ \\

HomeBench~\cite{li2025homebench}
& \cmark & \cmark & \cmark & \cmark & \cmark & \xmark & 170K+ \\

\textbf{Smart Home Instruct-Bench}
& \cmark & \xmark & \cmark & \xmark & \cmark & \cmark & 200 \\
\bottomrule
\end{tabular*}

\begin{tablenotes}[flushleft]
\footnotesize
\item[] VS/IS/VM/IM/MM follow the HomeBench taxonomy~\cite{li2025homebench};
the II column and the Smart Home Instruct-Bench row are added in this paper.
II denotes indirect instruction. For Smart Home Instruct-Bench, Size denotes
the 200-entry benchmark; the broader Smart Home Instruct dataset family also
includes Smart Home Instruct-SFT and Smart Home Instruct-Contrast.
\emph{Note:} \cmark{} = explicitly covered by natural-language instructions;
\xmark{} = not covered; \pxmark{} = partially or indirectly covered.
\end{tablenotes}
\end{threeparttable}
\end{table}
Table~\ref{tab:homebench_taxonomy_comparison} summarizes representative smart-home instruction resources. IFTTT contains more than 50K trigger-action recipes from the If-This-Then-That platform, capturing automated IoT behavior rules rather than direct natural-language command generation~\cite{yu2021ifttt}. SAGE evaluates autonomous LLM agents on 50 grounded smart-home execution tasks~\cite{rivkin2025aiotSmartHome}, while Home Assistant Requests provides assistant-style request-response pairs for controlling Home Assistant instances~\cite{acon96homeassistant2024}. These resources cover important aspects of smart-home automation and agent execution, but they provide limited emphasis on indirect intent commands, where users imply the desired device action through contextual or goal-oriented expressions.
In practical smart-home scenarios, command difficulty is not determined only by the number of devices or actions involved, but also by how user intent is expressed. Larger models can often recover from lexical variation, syntax errors, or indirect phrasing through stronger contextual reasoning. In contrast, compact models deployed under edge IoT constraints are more sensitive to intent ambiguity, which can lead to incorrect structured command generation. To study how architecture affects this setting, we evaluate three sub 0.2B-scale architectures under the same many-to-one structured command-output formulation, including the CMC encoder--decoder model~\cite{chen2023chatlmmini} and two self-trained encoder-only~\cite{hu2026encoderchineseslm} and decoder-only~\cite{hu2026decoderchineseslm} variants.
\begin{figure*}[!t]
\centering
\includegraphics[width=\textwidth]{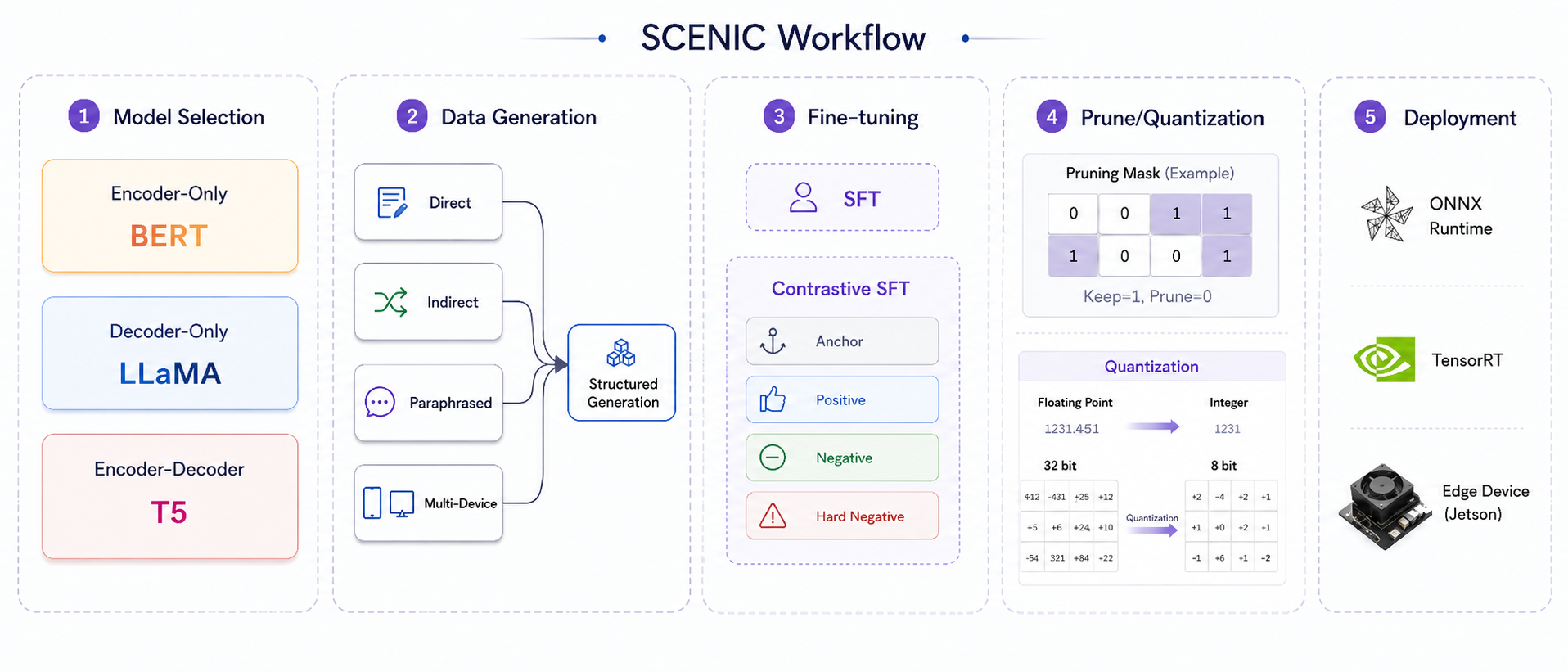}
\caption{Overview of the SCENIC framework.}
\label{fig:scenic_framework}
\end{figure*}
\subsection{Contrastive Learning}
Contrastive learning has been widely studied as a mechanism for improving representation structure. Khosla \textit{et al.}~\cite{khosla2020supervised} introduced supervised contrastive learning, where samples from the same class are pulled closer in the embedding space while samples from different classes are separated. This formulation improves discriminative representation learning beyond standard cross entropy training, particularly when different input variants correspond to the same semantic label. In text generation, Su \textit{et al.}~\cite{su2022contrastiveTextGeneration} showed that contrastive objectives can help calibrate the representation space of neural language models, reducing degeneration and improving generation coherence. More recently, Yan \textit{et al.}~\cite{yan2024contrastiveInstruction} proposed contrastive instruction tuning, which aligns semantically equivalent instruction instance pairs and improves robustness to unseen instruction variations. Unlike general contrastive instruction tuning, which improves instruction-following robustness by aligning semantically equivalent instruction--instance pairs~\cite{yan2024contrastiveInstruction}, SCENIC defines semantic equivalence as the same command: two smart-home commands are treated as compatible only when they map to the same executable structured IoT command output. To support this formulation, SCENIC constructs schema-aware contrastive tuples containing same-output positives, valid hard negatives with different structured outputs, and invalid negatives corresponding to unsupported device--action requests. This design encourages compact edge-oriented models to align paraphrased commands while separating superficially similar commands that require different or non-executable outputs. In SCENIC, contrastive SFT is used as an auxiliary training strategy for comparing representation behavior across architectures, rather than as the sole optimization objective.

\section{Methodology}
\label{sec:methodology}
This section presents the full SCENIC workflow. SCENIC is designed to generate and evaluate language models for edge-deployable many-to-one structured IoT command strings. For the remainder of the paper, SCENIC refers to the framework, Smart Home Instruct refers to the dataset family, Smart Home Instruct-SFT and Smart Home Instruct-Contrast are the training views, and Smart Home Instruct-Bench is the 200-entry benchmark.
As shown in Fig.~\ref{fig:scenic_framework}, SCENIC contains five stages: architecture selection, many-to-one data generation, triplet-loss contrastive SFT (C-SFT), pruning and quantization, and edge deployment. The consolidated implementation and experimental artifacts are provided in the SCENIC repository~\cite{hu2026scenic}.

\subsection{Architecture Selection}
\label{subsec:controlled_architectures}

SCENIC compares three similar corpus pretrained backbones under the same downstream
data and evaluation controls; the difference is the command-output interface.
The home action inventory is closed, so generalization is measured over unseen
linguistic forms and compositions within the fixed schema rather than over an
open action ontology. Let \(c_i=t(y_i)\) be the canonical command string for
instruction \(x_i\), let \(c_{i,u}\) denote its \(u\)-th target token, and let
\(\mathcal{T}_{\mathrm{train}}=\{t_1,\ldots,t_K\}\) be the deduplicated
training target inventory. For the encoder-only model, \(q_k=(w_k,b_k)\) are
the class parameters for target \(t_k\), \(r_{\mathrm{enc}}(x)\) is a pooled
instruction representation, and
\begin{equation}
\begin{aligned}
p_{\theta}(t_k\mid x)
&=
\frac{\exp(w_k^{\top}r_{\mathrm{enc}}(x)+b_k)}
{\sum_{j=1}^{K}\exp(w_j^{\top}r_{\mathrm{enc}}(x)+b_j)},\\
\mathcal{L}_{\mathrm{enc}}^{(i)}
&=-\log p_{\theta}(c_i\mid x_i).
\end{aligned}
\label{eq:encoder_loss}
\end{equation}
Decoder-only and encoder--decoder backbones are trained by token-level negative
log likelihood over the same canonical command string:
\begin{equation}
\begin{aligned}
\mathcal{L}_{\mathrm{dec}}^{(i)}
&=-\sum_{u=1}^{|c_i|}
\log p_{\theta}(c_{i,u}\mid x_i,c_{i,<u}),\\
\mathcal{L}_{\mathrm{encdec}}^{(i)}
&=-\sum_{u=1}^{|c_i|}
\log p_{\theta}(c_{i,u}\mid \mathrm{Enc}_{\theta}(x_i),c_{i,<u}).
\end{aligned}
\label{eq:generative_losses}
\end{equation}

\begin{table*}[!t]
\centering
\caption{Compact architecture and data-control summary.}
\label{tab:architecture_compact}
\footnotesize
\setlength{\tabcolsep}{3pt}
\renewcommand{\arraystretch}{1.18}
\begin{tabularx}{\textwidth}{@{}p{0.18\textwidth}XXX@{}}
\toprule
\textbf{Specification} &
\textbf{Encoder-Chinese-SLM~\cite{hu2026encoderchineseslm} } &
\textbf{Decoder-Chinese-SLM~\cite{hu2026decoderchineseslm} } &
\textbf{Encoder--Decoder} \\
\midrule
Model source &
Trained by the authors in this work &
Trained by the authors in this work &
Public compact encoder--decoder checkpoint \\
Architecture type &
Encoder-only &
Decoder-only &
Encoder--decoder \\
Backbone family &
BERT/RoBERTa-style bidirectional encoder &
LLaMA-style causal decoder &
T5-style conditional sequence-to-sequence model \\
Output formulation &
Bidirectional command representation and output selection &
Autoregressive structured-output generation &
Source-conditioned structured-output generation \\
Parameter scale &
$\sim$0.194B &
$\sim$0.196B &
$\sim$0.188B \\
Role in this work &
Representation baseline &
Causal generation baseline &
Conditional generation baseline \\
Pretraining status &
Author-trained compact Chinese encoder &
Author-trained compact Chinese decoder &
Public pretrained compact Chinese encoder--decoder \\
Tokenizer and vocabulary control &
\multicolumn{3}{p{0.78\textwidth}@{}}{
The author-trained encoder-only and decoder-only models use the same tokenized Chinese corpus recipe; all downstream comparisons use fixed tokenizer settings and the same Smart Home Instruct views to reduce data-induced confounding across architectures.
} \\
Downstream control &
\multicolumn{3}{p{0.78\textwidth}@{}}{
All models are fine-tuned, pruned, and evaluated using the same Smart Home Instruct-SFT, Smart Home Instruct-Contrast, Smart Home Instruct-Bench, structured exact-match metrics, and pruning/evaluation protocol. The comparison is a practical checkpoint-level comparison under shared downstream controls, not a pure pretraining-isolation study.
} \\
Implementation details &
\multicolumn{3}{p{0.78\textwidth}@{}}{
Training scripts, tokenizer settings, model configuration files, and repository-level implementation details for the author-trained checkpoints are provided in the cited code repositories~\cite{hu2026decoderchineseslm,hu2026encoderchineseslm}.
} \\
\bottomrule
\end{tabularx}
\end{table*}

\subsection{Smart Home Instruct Generation}
\label{subsec:smart-home-instruct}

SCENIC constructs Smart Home Instruct from a finite executable home-action
ontology. Let \(\mathscr{L}\) denote locations, \(\mathscr{D}\) devices,
\(\mathscr{S}(d)\) the finite set of services or actions supported by device
\(d\), and \(\Gamma\) argument keys. Each argument key
\(\gamma\in\Gamma\) has an admissible value set \(\mathscr{V}_{\gamma}\) and a
deterministic value normalizer \(\eta_{\gamma}\). A single action is
represented as
\begin{equation}
a_r=(\ell_r,d_r,s_r,\alpha_r),
\label{eq:action_schema}
\end{equation}
where \(\ell_r\in\mathscr{L}\cup\{\varnothing\}\), \(d_r\in\mathscr{D}\),
\(s_r\in\mathscr{S}(d_r)\), and
\(\alpha_r=\{(\gamma,v_{\gamma}):\gamma\in\Gamma_r,
v_{\gamma}\in\mathscr{V}_{\gamma}\}\) is an optional argument set. A
multi-action output is
\begin{equation}
y=(a_1,\ldots,a_R), \qquad R\geq 1.
\label{eq:structured_command}
\end{equation}
Let \(\mathscr{Y}_{\mathrm{valid}}\) denote the set of all multi-action outputs
whose action components satisfy these ontology constraints.

The canonical slot order is fixed everywhere as location, device,
service/action, and optional arguments. To define the normalization map, SCENIC
fixes total orders \(<_{\mathscr{L}}\), \(<_{\mathscr{D}}\),
\(<_{\mathscr{S}}\), and \(<_{\Gamma}\). Argument keys are sorted by
\(<_{\Gamma}\), values are normalized by \(\eta_{\gamma}\), and the resulting
argument list is denoted \(\alpha_r^{\uparrow}\). Actions are ordered by
\(\prec_{\nu}\), the lexicographic order over
\((\ell_r,d_r,s_r,\alpha_r^{\uparrow})\) induced by these total orders:
\begin{equation}
a_r\prec_{\nu} a_q
\Longleftrightarrow
(\ell_r,d_r,s_r,\alpha_r^{\uparrow})
<_{\mathrm{lex}}
(\ell_q,d_q,s_q,\alpha_q^{\uparrow}).
\label{eq:action_order}
\end{equation}
The normalized structured output is
\begin{equation}
\bar{y}=\nu(y),
\label{eq:command_normalization}
\end{equation}
where \(\nu(\cdot)\) sorts actions by \(\prec_{\nu}\), uses the fixed slot
order, serializes unspecified slots as \(\varnothing\), normalizes whitespace,
and applies deterministic argument formatting. The structured natural-language
rendering is
\begin{equation}
t(y):=\tau(\nu(y)).
\label{eq:canonical_rendering}
\end{equation}
The rendering map \(\tau\) is injective on normalized valid outputs and is paired
with a deterministic parser \(\pi\) such that
\begin{equation}
\pi(t(y))=\nu(y),\qquad \forall y\in\mathscr{Y}_{\mathrm{valid}}.
\label{eq:parser_inverse}
\end{equation}
Consequently, for any two valid outputs,
\begin{equation}
t(y_1)=t(y_2)
\Longleftrightarrow
\nu(y_1)=\nu(y_2).
\label{eq:canonical_equivalence}
\end{equation}
Thus, SCENIC uses natural-language output, but not open-ended dialogue: the
structured command string is a deterministic rendering of the structured command.

\begin{figure}[!t]
\centering
\includegraphics[width=0.92\columnwidth]{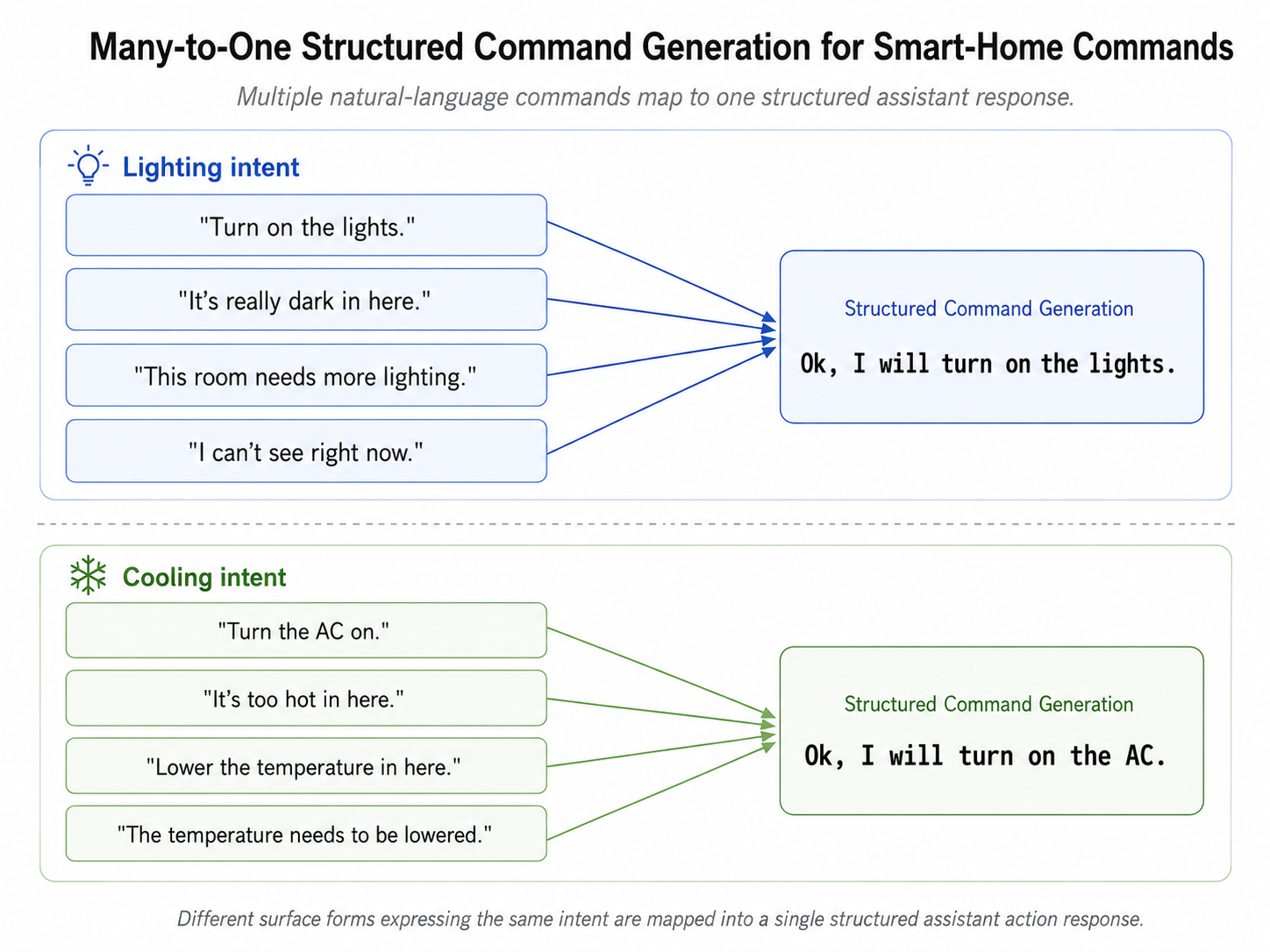}
\caption{Many-to-one Smart Home Instruct mapping.}
\label{fig:many_to_one}
\end{figure}

For each valid structured command output \(y\), SCENIC generates a variable-size
instruction set
\begin{equation}
X_y=\{x_{y,j}\}_{j=1}^{n_y},
\label{eq:instruction_variants}
\end{equation}
where the instructions include direct commands, indirect commands, and
paraphrased variants. Fig.~\ref{fig:many_to_one} illustrates the many-to-one
mapping from different surface instructions to a shared canonical command
string. The three Smart Home Instruct views are
\begin{equation}
\mathcal{E}_{\mathrm{SFT}}
=
\{(x_i,c_i)\}_{i=1}^{N_{\mathrm{SFT}}},
\qquad
c_i=t(y_i),
\label{eq:sft_dataset}
\end{equation}
\begin{equation}
\mathcal{E}_{\mathrm{Contrast}}
=
\{(x_i,x_i^{+},x_i^{-,\mathrm{hard}},
x_i^{-,\mathrm{inv}},c_i,c_i^{-})\}_{i=1}^{N_{\mathrm{C}}},
\label{eq:contrast_dataset}
\end{equation}
and
\begin{equation}
\mathcal{E}_{\mathrm{Bench}}
=
\{(x_j,c_j)\}_{j=1}^{200}.
\label{eq:bench_dataset}
\end{equation}
Smart Home Instruct-SFT is used for supervised task adaptation, Smart Home
Instruct-Contrast is used only for the triplet-loss C-SFT variant, and Smart
Home Instruct-Bench is the benchmark view used for reporting.
Duplicate normalized strings are removed when constructing the closed training
inventory
\begin{equation}
\mathcal{T}_{\mathrm{train}}
=
\{c_i:(x_i,c_i)\in\mathcal{E}_{\mathrm{SFT}}\}.
\label{eq:train_output_inventory}
\end{equation}
The benchmark target-overlap statistic is
\begin{equation}
\begin{aligned}
\rho_{\mathrm{overlap}}
&=
\frac{1}{200}\sum_{j=1}^{200}
\mathbf{1}\!\left\{c_j\in\mathcal{T}_{\mathrm{train}}\right\},\\
\rho_{\mathrm{OOV}}
&=
1-\rho_{\mathrm{overlap}}.
\end{aligned}
\label{eq:target_overlap}
\end{equation}
For the current Smart Home Instruct artifacts,
\(\rho_{\mathrm{overlap}}=200/200=1.00\), so all benchmark target strings are
covered by the closed training inventory.

Evaluation uses normalized exact match at top \(K\):
\begin{equation}
\mathrm{EM@}K
=
\frac{1}{N}\sum_{i=1}^{N}
\mathbf{1}\{c_i\in\operatorname{TopK}(x_i)\}.
\label{eq:emk}
\end{equation}
For encoder-only evaluation, \(\operatorname{TopK}(x_i)\) is ranked over
\(\mathcal{T}_{\mathrm{train}}\) by classifier score. For decoder-only and
encoder--decoder evaluation, beam size is 5, the maximum decoding budget is 128
new tokens, decoded strings are normalized through the same
\(\nu(\cdot)\)-and-\(\tau(\cdot)\) pipeline, duplicates are removed in
beam-score order, and the first \(K\) remaining candidates are used. Wilson
score intervals~\cite{wilson1927probable} are reported for representative
Smart Home Instruct-Bench EM@1 rows.
\begin{table}[!htbp]
\centering
\caption{Smart Home Instruct data views used in SCENIC.}
\label{tab:data_views}
\scriptsize
\setlength{\tabcolsep}{3pt}
\renewcommand{\arraystretch}{1.18}
\begin{tabularx}{\columnwidth}{p{0.34\columnwidth}p{0.21\columnwidth}X}
\toprule
\textbf{View} & \textbf{Size} & \textbf{Role} \\
\midrule
Smart Home Instruct-SFT & 9,772 pairs & Supervised adaptation and pruning calibration. \\
Smart Home Instruct-Contrast & 9,772 tuples & Anchor, same-target positive, valid hard negative, and invalid negative tuples. \\
Smart Home Instruct-Bench & 200 pairs & Benchmark view with 70 easy, 65 medium, and 65 hard examples. \\
Unique canonical targets & 1,411 strings & Deduplicated target inventory in the SFT and Contrast artifacts. \\
\bottomrule
\end{tabularx}
\end{table}

\subsection{Command-Output-Level Triplet Loss}
\label{subsec:csft}

Smart Home Instruct-Contrast supplies anchor--positive--negative tuples for a
triplet-loss auxiliary objective; Smart Home Instruct-Bench contains no negative
examples. Each record is organized as
\begin{equation}
\xi_i =
\bigl(x_i,x_i^{+},x_i^{-},\tilde{x}_i^{-},c_i,c_i^{-}\bigr),
\label{eq:csft_tuple}
\end{equation}
where \(x_i^{+}\) shares the target \(c_i\), \(x_i^{-}\) is a valid hard
negative whose target \(c_i^{-}\neq c_i\), and \(\tilde{x}_i^{-}\) is an
invalid negative that violates the device--action schema. Let
\(\bar{y}_i=\nu(y_i)\), and let \(\operatorname{Dev}(\bar{y})\) and
\(\operatorname{Svc}(\bar{y})\) denote the devices and services/actions in a
normalized output. Anchor-conditioned candidate sets are
\begin{equation}
\begin{aligned}
\mathcal{P}_i
&=\{x_j:j\neq i,\ c_j=c_i\},\\
\mathcal{N}_i^{\mathrm{hard}}
&=\{x_j:c_j\neq c_i,\ 
\operatorname{Dev}(\bar{y}_j)\cap\operatorname{Dev}(\bar{y}_i)\neq\varnothing,\\
&\qquad
\operatorname{Svc}(\bar{y}_j)\cap\operatorname{Svc}(\bar{y}_i)\neq\varnothing\},\\
\mathcal{N}_i^{\mathrm{inv}}
&=\{\tilde{x}_i^{-}:\tilde{x}_i^{-}\ \text{is schema-invalid for }x_i\}.
\end{aligned}
\label{eq:contrastive_sampling_sets}
\end{equation}
This makes hard negatives anchor-conditioned rather than the full
different-target complement. If \(\mathcal{P}_i=\varnothing\), SCENIC first
creates a same-output paraphrase; if no positive can be produced, the anchor
remains in Smart Home Instruct-SFT but is omitted from the auxiliary triplet
term.

\begin{figure}[!t]
\centering
\includegraphics[width=0.92\columnwidth]{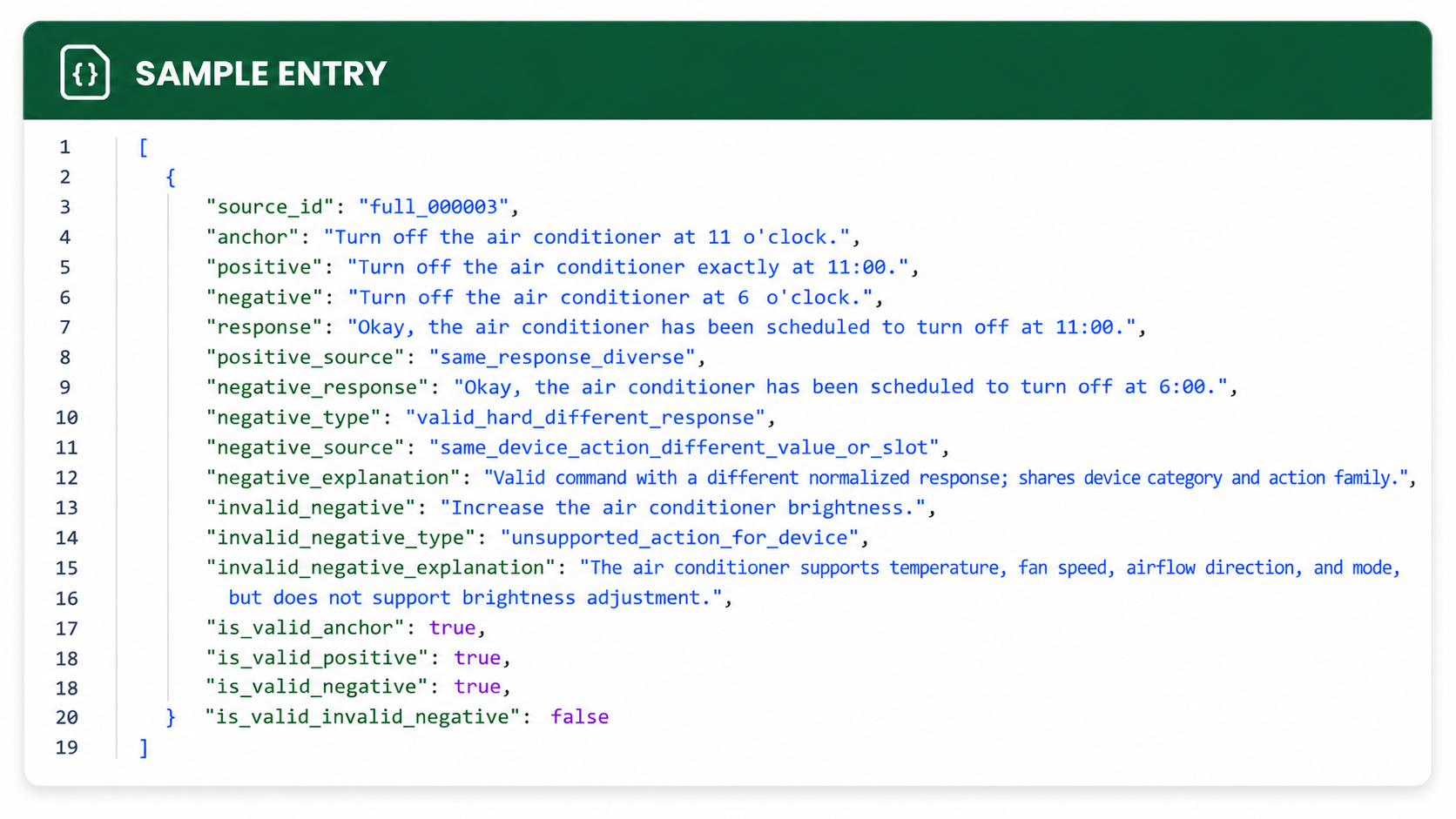}
\caption{Translated Smart Home Instruct-Contrast entry.}
\label{fig:dataset_example}
\end{figure}

The released C-SFT training code loads one negative field per run. The reported
C-SFT rows use the default valid hard negative field \(x_i^{-}\); the
\(\tilde{x}_i^{-}\) field is retained for schema validation and optional
invalid-negative runs, but is not used simultaneously in the reported loss.

\begin{algorithm}[!t]
\caption{Triplet-loss contrastive SFT used in SCENIC.}
\label{alg:contrastive_sft}
\footnotesize
\begin{algorithmic}[1]
\REQUIRE Batch \(\mathcal{B}=\{(x_i,x_i^{+},x_i^{-},c_i)\}\), backbone \(A\), margin \(m\), weight \(\lambda_{\mathrm{tri}}\).
\ENSURE Updated model parameters.
\FOR{each tuple \(i\in\mathcal{B}\)}
\STATE Compute the supervised task loss for anchor and positive inputs.
\STATE Encode \(x_i\), \(x_i^{+}\), and \(x_i^{-}\); mean-pool and normalize the representations.
\STATE Compute \(\mathcal{L}_{\mathrm{tri}}^{(i)}=[m+d(r_i,r_i^{+})-d(r_i,r_i^{-})]_{+}\).
\STATE Combine \(\mathcal{L}^{(i)}=\mathcal{L}_{\mathrm{task}}^{(i)}+\lambda_{\mathrm{tri}}\mathcal{L}_{\mathrm{tri}}^{(i)}\).
\ENDFOR
\STATE Update \(A\) using the batch mean loss.
\end{algorithmic}
\end{algorithm}

For a backbone \(A\), the representation extractor is
\begin{equation}
r_A(x)=
\frac{\operatorname{Pool}_A(H_A(x))}
{\|\operatorname{Pool}_A(H_A(x))\|_2+\epsilon},
\label{eq:representation_extractor}
\end{equation}
where the released encoder--decoder C-SFT implementation uses attention-mask mean pooling
over encoder hidden states. With \(d(u,v)=1-u^{\top}v\) for normalized vectors,
the implemented triplet term is
\begin{equation}
\mathcal{L}_{\mathrm{tri}}^{(i)}
=
\left[
m+d\!\left(r_A(x_i),r_A(x_i^{+})\right)
-d\!\left(r_A(x_i),r_A(x_i^{-})\right)
\right]_{+}.
\label{eq:triplet_loss}
\end{equation}
The admissible margin range is \(0<m<2\); the reported C-SFT setting uses
\(m=0.5\) and \(\lambda_{\mathrm{tri}}=0.1\). For the generative encoder--decoder script, the
task term averages anchor and positive sequence losses,
\begin{equation}
\mathcal{L}_{\mathrm{gen}}^{(i)}
=\tfrac{1}{2}\left[
\mathcal{L}_{\mathrm{encdec}}(x_i,c_i)+
\mathcal{L}_{\mathrm{encdec}}(x_i^{+},c_i)
\right],
\label{eq:csft_gen_loss}
\end{equation}
and the reported triplet-loss C-SFT objective is
\begin{equation}
\mathcal{L}_{\mathrm{C\mbox{-}SFT}}
=
\frac{1}{|\mathcal{B}|}\sum_{i\in\mathcal{B}}
\left(\mathcal{L}_{\mathrm{task}}^{(i)}
+\lambda_{\mathrm{tri}}\mathcal{L}_{\mathrm{tri}}^{(i)}\right),
\label{eq:csft_objective}
\end{equation}
where \(\mathcal{L}_{\mathrm{task}}\) is the architecture-specific supervised
loss in Eqs.~\eqref{eq:encoder_loss}--\eqref{eq:generative_losses}; for the
released encoder--decoder run it is \(\mathcal{L}_{\mathrm{gen}}\) in
Eq.~\eqref{eq:csft_gen_loss}.

\begin{table*}[t]
\centering
\caption{Exploratory Smart Home Instruct-Bench EM@1/EM@5 results.}
\label{tab:scenic_main_results}
\scriptsize
\setlength{\tabcolsep}{2.45pt}
\renewcommand{\arraystretch}{1.03}

\begin{threeparttable}
\begin{tabular}{@{}lllrrrrrr@{}}
\toprule
\textbf{Arch.} &
\textbf{Train.} &
\textbf{Setting} &
\textbf{Train} &
\textbf{Bench.} &
\textbf{Easy} &
\textbf{Med.} &
\textbf{Hard} &
\textbf{Method} \\
\midrule

Enc-only & Base & Unadapted
& 2.30/7.10 & \textbf{1.50/7.00} & \textbf{2.86/10.00} & \textbf{1.54/10.77} & 0.00/0.00 & Base model \\
Dec-only & Base & Unadapted
& 0.00/0.00 & 0.00/0.00 & 0.00/0.00 & 0.00/0.00 & 0.00/0.00 & Base model \\
Enc-Dec & Base & Unadapted
& 0.00/0.00 & 0.00/0.00 & 0.00/0.00 & 0.00/0.00 & 0.00/0.00 & Base model \\

\midrule
Enc-only & SFT & Dense
& 80.19/96.38 & 85.50/98.50 & 65.71/95.71 & 100.00/100.00 & 92.31/100.00 & Dense baseline \\
Dec-only & SFT & Dense
& \textbf{99.95/100.00} & \textbf{99.00/99.50} & \textbf{97.14/98.57} & \textbf{100.00/100.00} & \textbf{100.00/100.00} & Dense baseline \\
Enc-Dec & SFT & Dense
& 91.18/98.47 & 95.00/99.50 & 85.71/98.57 & \textbf{100.00/100.00} & \textbf{100.00/100.00} & Dense baseline \\

\midrule
Enc-only & SFT & 30\%
& 39.80/64.60 & 36.00/85.00 & 28.57/74.29 & 61.54/92.31 & 18.46/89.23 & Prog. mag. \\
Dec-only & SFT & 30\%
& \textbf{98.82/99.88} & \textbf{99.50/100.00} & \textbf{98.57/100.00} & \textbf{100.00/100.00} & \textbf{100.00/100.00} & Magnitude \\
Enc-Dec & SFT & 30\%
& 90.97/98.43 & 96.50/99.50 & 90.00/98.57 & \textbf{100.00/100.00} & \textbf{100.00/100.00} & Gradient \\

\midrule
Enc-only & SFT & 50\%
& 35.20/64.02 & 35.50/86.50 & 22.86/68.57 & 58.46/100.00 & 26.15/92.31 & Prog. mag. \\
Dec-only & SFT & 50\%
& 63.07/77.55 & 32.00/44.00 & 85.71/97.14 & 0.00/20.00 & 6.15/10.77 & Gradient \\
Enc-Dec & SFT & 50\%
& \textbf{86.74/96.95} & \textbf{97.00/99.50} & \textbf{91.43/98.57} & \textbf{100.00/100.00} & \textbf{100.00/100.00} & WANDA \\

\midrule
Enc-only & C-SFT & Dense
& 67.94/90.87 & 78.50/95.50 & 55.71/87.14 & 96.92/100.00 & 84.62/100.00 & Dense baseline \\
Dec-only & C-SFT & Dense
& \textbf{100.00/100.00} & 97.00/97.50 & 91.43/92.86 & \textbf{100.00/100.00} & \textbf{100.00/100.00} & Dense baseline \\
Enc-Dec & C-SFT & Dense
& 95.65/97.96 & \textbf{98.00/98.50} & \textbf{94.29/95.71} & \textbf{100.00/100.00} & \textbf{100.00/100.00} & Dense baseline \\

\midrule
Enc-only & C-SFT & 30\%
& 42.57/63.93 & 56.50/85.50 & 35.71/78.57 & 72.31/95.38 & 63.08/83.08 & Prog. mag. \\
Dec-only & C-SFT & 30\%
& \textbf{97.85/99.65} & \textbf{99.50/100.00} & \textbf{98.57/100.00} & \textbf{100.00/100.00} & \textbf{100.00/100.00} & Magnitude \\
Enc-Dec & C-SFT & 30\%
& 94.81/97.26 & 98.00/99.00 & 94.29/97.14 & \textbf{100.00/100.00} & \textbf{100.00/100.00} & Gradient \\

\midrule
Enc-only & C-SFT & 50\%
& 28.28/55.58 & 28.00/77.50 & 17.14/68.57 & 50.77/98.46 & 16.92/66.15 & Prog. mag. \\
Dec-only & C-SFT & 50\%
& 18.55/40.88 & 16.00/27.00 & 45.71/77.14 & 0.00/0.00 & 0.00/0.00 & Gradient \\
Enc-Dec & C-SFT & 50\%
& \textbf{90.67/95.82} & \textbf{95.50/99.50} & \textbf{91.43/100.00} & \textbf{100.00/100.00} & \textbf{95.38/98.46} & Gradient \\

\bottomrule
\end{tabular}

\begin{tablenotes}[flushleft]
\footnotesize
\item[] \textit{Note:} Each metric entry reports EM@1/EM@5 in percent.
C-SFT denotes supervised fine-tuning with the triplet-loss auxiliary objective.
Dense denotes the unpruned checkpoint. Bold values mark the highest value
within each comparable training and sparsity block.
\end{tablenotes}
\end{threeparttable}
\end{table*}
\subsection{Edge-Compression Evaluation Protocol}
\label{subsec:pruning}

After SFT and C-SFT, SCENIC treats pruning as an edge-compression stress test
over selected compact checkpoints. Pruning is applied to transformer linear
weights, calibrated only on training data, and reported as exploratory
benchmark-view analysis unless validation-based selection is available. The
evaluated methods include one-shot magnitude pruning~\cite{han2015learningWeightsConnections},
gradient-based importance pruning~\cite{molchanov2019importance},
WANDA~\cite{sun2023wanda}, NVIDIA 2:4 structured sparsity~\cite{mishra2021acceleratingSparse},
and progressive magnitude pruning~\cite{zhu2017prune}.

\subsection{ONNX Export and Deployment-Oriented Evaluation}
\label{subsec:onnx_eval}

SCENIC exports selected checkpoints to ONNX under FP16 and INT8 settings to test
deployment compatibility after reduced-precision conversion. The export protocol
records stored model size, latency, throughput, and EM@1/EM@5 under fixed
decoding settings, while hardware-aware acceleration is evaluated separately
through TensorRT encoder profiling.

\section{Experimental Results}
\label{sec:experimental_results}

This section evaluates SCENIC under Smart Home Instruct-SFT, Smart Home
Instruct-Contrast, pruning, ONNX export, and TensorRT-oriented profiling. The evaluation is organized in three stages. First, the three compact Chinese model architectures are compared under dense SFT and C-SFT. Second, the same architectures are evaluated after pruning at 30\% and 50\% target sparsity to measure sparse accuracy retention. Third, the selected encoder--decoder deployment path is exported to ONNX under FP16 and INT8 runtime settings, while TensorRT encoder profiling is reported separately as a hardware-aware acceleration probe.
All rows use the EM@1/EM@5 protocol defined in Sec.~\ref{subsec:smart-home-instruct}. The pruning rows are reported as exploratory benchmark-view results rather than locked validation-selected deployment choices.
\begin{table}[!h]
\centering
\caption{Smart Home Instruct-Bench difficulty breakdown.}
\label{tab:benchmark_difficulty_results}
\scriptsize
\setlength{\tabcolsep}{3pt}
\renewcommand{\arraystretch}{1.16}
\begin{tabularx}{\columnwidth}{@{}L{0.23\columnwidth}YYY@{}}
\toprule
\textbf{Criterion} & \textbf{Easy} & \textbf{Medium} & \textbf{Hard} \\
\midrule
Examples & 70 & 65 & 65 \\
Device scope & Single-device & Multi-device & Multi-device \\
Command actions & 1 & 2 & $\geq 3$ \\
Prompt type & Direct/indirect & Two-action command & Scenario-indirect \\
Evaluation focus & Basic command & Multi-device execution & Scenario-level action grounding \\
\bottomrule
\end{tabularx}
\end{table}

\begin{table}[!h]
\centering
\caption{Representative 95\% Wilson confidence intervals on Smart Home Instruct-Bench EM@1.}
\label{tab:benchmark_ci}
\scriptsize
\setlength{\tabcolsep}{3.2pt}
\renewcommand{\arraystretch}{1.16}
\begin{tabularx}{\columnwidth}{@{}Y C{0.16\columnwidth}C{0.27\columnwidth}@{}}
\toprule
\textbf{Configuration} & \textbf{EM@1} & \textbf{95\% CI} \\
\midrule
Dec-only SFT dense & 99.0 & [96.4, 99.7] \\
Enc-Dec C-SFT dense & 98.0 & [95.0, 99.2] \\
Enc-Dec SFT 50\% WANDA & 97.0 & [93.6, 98.6] \\
Enc-Dec C-SFT 50\% gradient & 95.5 & [91.7, 97.6] \\
Enc-Dec pruned INT8 export & 91.0 & [86.2, 94.2] \\
\bottomrule
\end{tabularx}
\end{table}

\begin{table*}[!h]
\centering
\caption{ONNX runtime comparison at input length 64.}
\label{tab:onnx_latency_results}
\scriptsize
\setlength{\tabcolsep}{2.4pt}
\renewcommand{\arraystretch}{1.18}
\begin{tabular}{@{}C{0.085\textwidth}C{0.105\textwidth}C{0.090\textwidth}C{0.090\textwidth}C{0.095\textwidth}C{0.095\textwidth}C{0.070\textwidth}C{0.070\textwidth}C{0.120\textwidth}@{}}
\toprule
\textbf{Variant} &
\textbf{Precision} &
\shortstack{\textbf{Mean}\\\textbf{Lat.}} &
\shortstack{\textbf{P95}\\\textbf{Lat.}} &
\textbf{Throughput} &
\shortstack{\textbf{Model}\\\textbf{Size}} &
\textbf{EM@1} &
\textbf{EM@5} &
\textbf{Speedup} \\
 & & \textbf{(ms)} & \textbf{(ms)} & \textbf{(QPS)} & \textbf{(MB)} & \textbf{(\%)} & \textbf{(\%)} & \textbf{vs. FP16 Dense} \\
\midrule
Dense  & ONNX FP16 & 151.15 & 167.79 & 6.62 & 1056.21 & 97.00 & 98.50 & 1.00$\times$ \\
Pruned & ONNX FP16 & 137.04 & 157.54 & 7.30 & 1056.21 & 90.50 & 98.50 & 1.10$\times$ \\
Dense  & ONNX INT8 & 172.78 & 188.40 & 5.79 & 788.18  & 97.50 & 98.50 & 0.87$\times$ \\
Pruned & ONNX INT8 & 160.95 & 176.34 & 6.21 & 788.18  & 91.00 & 99.00 & 0.94$\times$ \\
\bottomrule
\end{tabular}
\end{table*}
\begin{table}[!h]
\centering
\caption{TensorRT encoder profiling on Jetson Orin.}
\label{tab:tensorrt_encoder_results}
\scriptsize
\setlength{\tabcolsep}{2.4pt}
\renewcommand{\arraystretch}{1.16}
\begin{adjustbox}{max width=\columnwidth,center}
\begin{tabular}{@{}C{0.16\columnwidth}C{0.18\columnwidth}C{0.18\columnwidth}C{0.22\columnwidth}C{0.18\columnwidth}@{}}
\toprule
\shortstack{\textbf{Seq.}\\\textbf{Len.}} &
\shortstack{\textbf{FP16}\\\textbf{Lat.}} &
\shortstack{\textbf{INT8}\\\textbf{Lat.}} &
\shortstack{\textbf{INT8}\\\textbf{TPS}} &
\shortstack{\textbf{INT8}\\\textbf{Gain}} \\
 & \textbf{(ms)} & \textbf{(ms)} & \textbf{(tok/s)} & \textbf{vs. FP16} \\
\midrule
1   & 5.21 & 3.63 & 275.35    & 1.43$\times$ \\
64  & 6.25 & 3.46 & 18,478.79 & 1.80$\times$ \\
128 & 6.49 & 6.75 & 18,957.94 & 0.96$\times$ \\
\bottomrule
\end{tabular}
\end{adjustbox}

\tabnoteskip
\footnotesize
\textit{Note:} Measurements cover encoder execution only and do not represent end-to-end encoder--decoder generation latency. The INT8 TensorRT encoder engine size is 120.90 MB.
\end{table}
\subsection{Benchmark Composition}
\label{subsec:benchmark_results}

SCENIC is evaluated on Smart Home Instruct-Bench, a 200-entry 
smart-home instruction benchmark. The benchmark contains easy, medium, and hard
instructions to test whether each model can preserve structured command accuracy
across direct single-device commands, two-action commands, and scenario-level
multi-action commands. Table~\ref{tab:benchmark_difficulty_results} summarizes
the benchmark composition.

\subsection{Dense and Pruned Accuracy Results}
\label{subsec:pruned_results}

Table~\ref{tab:scenic_main_results} reports the base models, dense baselines,
and representative exploratory pruning rows used to analyze compression
robustness under this experimental configuration.

Because Smart Home Instruct-Bench contains 200 examples, one example
corresponds to 0.5 percentage points; small differences are therefore
interpreted cautiously, and the analysis emphasizes broad architecture and
compression trends rather than statistical significance.
Table~\ref{tab:benchmark_ci} reports representative Wilson intervals for key EM@1 rows.

The unadapted base models provide limited command-generation ability, while
all three architectures learn the canonical command task after supervision. In
dense SFT, the decoder-only model obtains the strongest observed Smart Home
Instruct-Bench row, reaching 99.00/99.50 EM@1/EM@5; under C-SFT, the
encoder--decoder model reaches the strongest dense C-SFT row, 98.00/98.50.

At 30\% sparsity, the best decoder-only magnitude-pruning row remains strong
under both SFT and C-SFT, reaching 99.50/100.00 EM@1/EM@5. At 50\% sparsity,
the same architecture becomes unstable, falling to 32.00/44.00 under SFT and
16.00/27.00 under C-SFT, with most degradation appearing in medium and hard
multi-action commands.

The encoder-only model remains pruning-sensitive even with progressive
magnitude pruning, whereas the encoder--decoder model is the most stable
high-sparsity candidate. At 50\% sparsity, the encoder--decoder SFT row with
WANDA reaches 97.00/99.50 EM@1/EM@5, and the encoder--decoder C-SFT row with
gradient pruning reaches 95.50/99.50.

C-SFT is therefore interpreted as a triplet-loss auxiliary variant rather than a
uniformly improving recipe: it supports strong encoder--decoder dense behavior
and some encoder-only moderate-sparsity gains, but it does not prevent
decoder-only collapse under aggressive pruning.

\subsection{ONNX Runtime Results}
\label{subsec:onnx_runtime_results}

Table~\ref{tab:onnx_latency_results} reports ONNX FP16 and INT8 runtime results
for dense and pruned encoder--decoder exports at input length 64. The pruned
INT8 export preserves 91.00/99.00 EM@1/EM@5 and reduces stored model size, but
ONNX runtime latency does not improve over the dense FP16 reference on this
software stack.

\subsection{TensorRT Encoder Profiling Results}
\label{subsec:tensorrt_encoder_results}

To examine hardware-aware acceleration potential, the CMC encoder component is profiled under TensorRT FP16 and INT8 execution on Jetson Orin using the NVIDIA 2:4 sparse export path. This experiment measures encoder execution only and should not be interpreted as end-to-end encoder--decoder generation latency. Table~\ref{tab:tensorrt_encoder_results} reports encoder latency, INT8 tokens per second, and INT8 speedup relative to FP16 at sequence lengths 1, 64, and 128.

At sequence length 64, INT8 TensorRT execution reduces encoder latency from 6.25 ms to 3.46 ms, corresponding to a 1.80$\times$ speedup. At sequence length 128, INT8 latency is 6.75 ms compared with 6.49 ms for FP16, showing that the acceleration benefit depends on sequence length and runtime support. These results indicate that hardware-specific kernels can accelerate selected model components, but full SCENIC deployment still requires decoder-side and end-to-end generation profiling.

\subsection{Summary of Findings}
\label{subsec:results_summary}

The experimental results support three main findings. First, all compact
backbones require SCENIC task supervision before they can perform structured 
smart-home command generation. Second, dense training fit and moderately sparse
accuracy do not reliably predict aggressive-compression robustness: the
decoder-only model provides the highest observed dense training fit and the
highest observed 30\% benchmark scores, but collapses at 50\% sparsity. Third, the
encoder--decoder architecture provides the most stable high-sparsity behavior and
remains the most suitable deployment candidate in this SCENIC evaluation. This
supports the main SCENIC motivation: architecture selection for edge
smart-home assistants should account for structured-output accuracy, pruning
robustness, and deployment export behavior together.

The deployment results further show that accuracy retention and stored model
size reduction do not guarantee runtime acceleration; hardware-specific speedup
depends on optimized kernels such as TensorRT, NPU runtimes, or ASIC-oriented
sparse execution.

\section{Conclusion}
This paper identifies compression-aware structured command generation as a key challenge for edge IoT systems. To address this is we proposed SCENIC, a framework that combines Smart Home Instruct data construction, compact architecture comparison, triplet-loss C-SFT, pruning, ONNX export, and TensorRT-oriented profiling. The results show that dense accuracy alone is insufficient for selecting edge IoT command models: decoder-only backbones achieve the strongest dense and moderate-sparsity rows, whereas the encoder–decoder deployment path is more stable under aggressive pruning. ONNX INT8 export reduces stored model size while preserving usable command accuracy, but runtime acceleration depends on hardware-specific kernels, as shown by the component-level TensorRT encoder results. Future work will extend SCENIC with locked validation-based pruning selection, larger prompt-held-out and composition-held-out benchmarks, and end-to-end edge latency and energy evaluation.

\section*{Acknowledgment}
The authors acknowledge the use of AI-assisted tools for manuscript editing,
LaTeX formatting, and revision support. All technical content, experimental
results, and conclusions were reviewed and verified by the authors, who take
full responsibility for the accuracy and integrity of this work.

\bibliographystyle{IEEEtran}
\bibliography{references}

\end{document}